\documentclass[journal]{IEEEtran}

\usepackage{times}
\usepackage{epsfig}
\usepackage{graphicx}
\usepackage{subfig}
\usepackage{amsmath}
\usepackage{amssymb}
\usepackage{bm}
\usepackage{multirow}
\usepackage{colortbl}
\usepackage{algorithm}
\usepackage[nospace,nocompress]{cite}

\usepackage{pifont}
\newcommand{\cmark}{\ding{51}}%
\newcommand{\xmark}{\ding{55}}%
\ifCLASSINFOpdf
\else
\fi
\begin{document}

\title{Scalable Deep Learning Logo Detection}

\author{\IEEEauthorblockN{Hang Su\IEEEauthorrefmark{1},
Shaogang Gong\IEEEauthorrefmark{2},
Xiatian Zhu\IEEEauthorrefmark{3}}\\
\IEEEauthorblockA{\IEEEauthorrefmark{1} \IEEEauthorrefmark{2} Queen Mary University of London}
\IEEEauthorblockA{\IEEEauthorrefmark{3} Vision Semantics Ltd.}
}

%



\maketitle

\begin{abstract}
Existing logo detection methods usually consider a small number of logo classes and 
limited images per class with a strong assumption of
requiring tedious object bounding box annotations,
therefore not scalable to real-world dynamic applications.
In this work, we tackle these challenges
by exploring the webly data learning principle without 
the need for exhaustive manual labelling.
Specifically, we propose a novel incremental learning approach, called Scalable Logo Self-co-Learning (SL$^2$), capable of
automatically self-discovering informative training images 
from noisy web data 
for progressively improving model capability
in a cross-model co-learning manner.
%
Moreover, we introduce a very large (2,190,757 images of 194 logo classes) logo dataset ``WebLogo-2M'' by 
an {automatic web data} collection and processing method.
Extensive comparative evaluations demonstrate the superiority 
of the proposed SL$^2$ method over
the state-of-the-art strongly and weakly supervised detection models
and contemporary webly data learning approaches.
\end{abstract}

\begin{IEEEkeywords}
Webly Learning, Scalable Logo Detection, Incremental Learning,
Self-Learning, Co-Learning.
\end{IEEEkeywords}

%
\IEEEpeerreviewmaketitle

\section{Introduction}
\label{sec:intro}

\begin{figure} 
	\centering
	\includegraphics[width=1.0\linewidth]{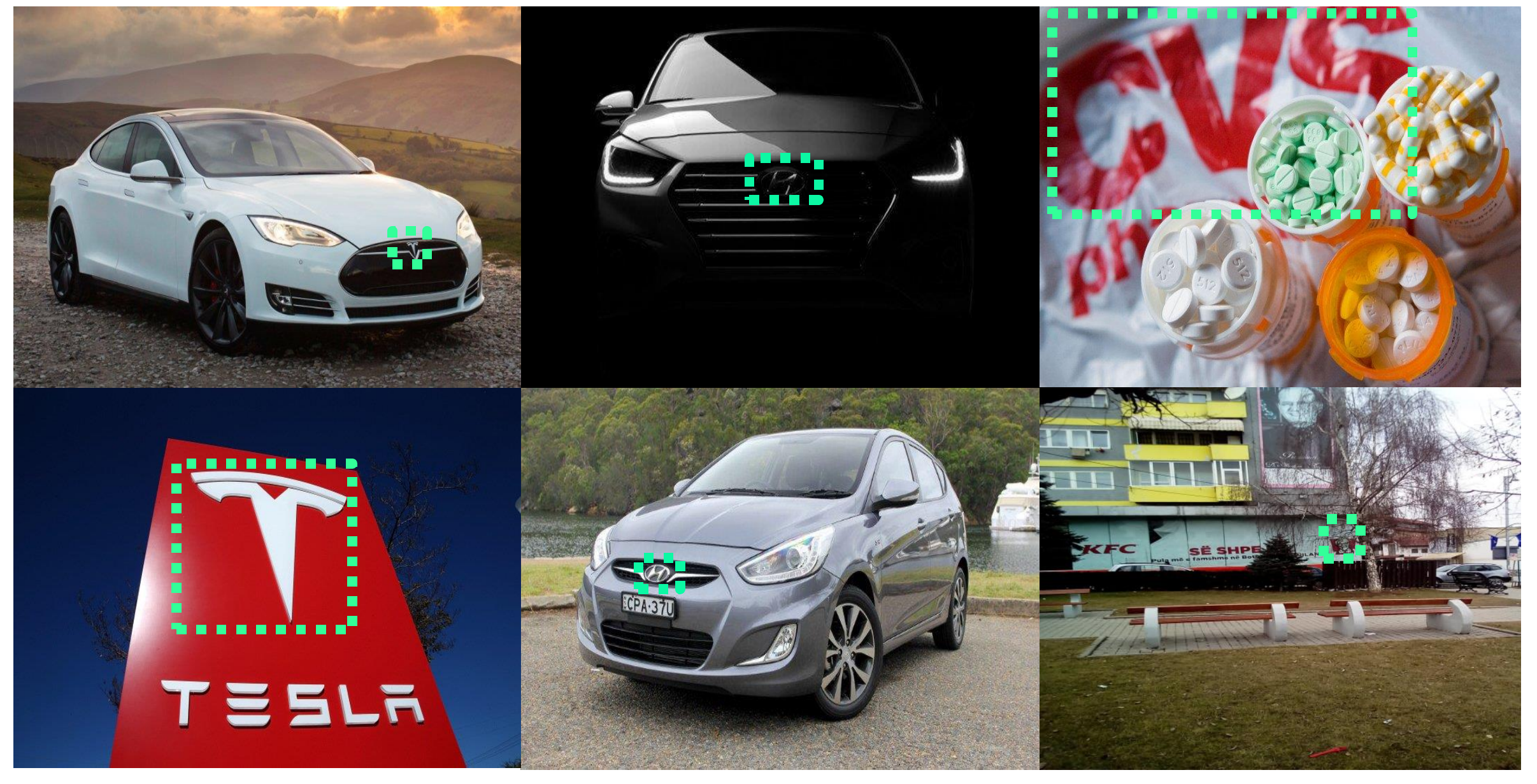}
	\vskip -0.0cm
	\caption{
	Logo detection challenges: significant logo variation in object size, illumination,
	background clutter, and occlusion.
	}
	\label{fig:challenges}
	\vspace{-0.5cm}
\end{figure}

Automated logo detection from unconstrained ``in-the-wild'' images
benefits a wide range of applications, 
e.g. 
brand trend prediction for commercial research and
vehicle logo recognition for intelligent transportation
\cite{romberg2011scalable,romberg2013bundle,pan2013vehicle}.
This is inherently a challenging task 
due to the presence of many logos  
in diverse context with uncontrolled illumination,
low-resolution, and background clutter (Fig. \ref{fig:challenges}).
%
%
%
%
%

Existing logo detection methods typically consider a small number of logo classes
with the need for large sized training data annotated at the logo object instance level, i.e. object bounding boxes
\cite{joly2009logo,kalantidis2011scalable,romberg2011scalable,revaud2012correlation,romberg2013bundle,boia2014local,li2014logo,pan2013vehicle}. 
Whilst this controlled setting allows for a straightforward adoption of 
the state-of-the-art object detection models \cite{ren2015faster,girshick2015fast,redmon2017yolo9000},
it is 
unscalable to real-world logo detection applications
when a much larger number of logo classes are of interest but limited
by (1) the extremely high cost for constructing 
large scale logo dataset with exhaustive logo
instance bounding box labelling therefore unavailability \cite{russakovsky2015imagenet}; 
and (2) lacking incremental model learning to progressively
update and expand the model to increasingly more training
data without fine-grained labelling. 
Existing models are mostly one-pass trained and statically
generalised to new test data.

%

\begin{table*}  
	\centering
	\setlength{\tabcolsep}{0.4cm}
	\caption{
		Statistics and characteristics of existing logo detection benchmarking datasets.
	}
	\vskip -0.0cm
	\label{tab:dataset}
	\begin{tabular}{c||c|c|c|c|c|c|c}
		\hline
		Dataset & Logo Classes & Images & Supervision & Noisy & Construction & Scalability & Availability \\
		\hline \hline
		TopLogo-10 \cite{su2016deep}
		& 10 & 700 & Object-Level & \xmark & Manually & Weak & \cmark \\
		\hline
		TennisLogo-20 \cite{liao2017mutual}
		& 20 & 2,000 & Object-Level & \xmark & Manually & Weak & \xmark \\
		\hline
		FlickrLogos-27 \cite{kalantidis2011scalable}
		& 27 & 810 & Object-Level  & \xmark & Manually  & Weak & \cmark \\
		\hline
		FlickrLogos-32 \cite{romberg2011scalable}
		& 32 & 2,240 & Object-Level & \xmark & Manually  & Weak & \cmark \\
		\hline
		Logo32-270 \cite{li2017graphic} 
		& 32 & 8,640 & Object-Level & \xmark & Manually  & Weak & \xmark
		\\ \hline
		BelgaLogos \cite{joly2009logo}  
		& 37  & 1321  & Object-Level & \xmark & Manually & Weak & \cmark \\
		\hline
		LOGO-NET \cite{hoi2015logo}
		& 160 & 73,414 & Object-Level & \xmark & Manually & Weak & \xmark \\
		\hline \hline
		\bf WebLogo-2M (Ours)
		&\bf 194 &\bf 2,190,757 & Image-Level & \cmark & Automatically & Strong & \cmark \\
		\hline
	\end{tabular}
\end{table*}

In this work, we consider scalable logo detection learning in a very large
collection of unconstrained images without 
exhaustive fine-grained object instance level labelling for model training.
Given that existing datasets mostly have small numbers of logo classes, 
one possible strategy is to learning from a small set of labelled
training classes and adopting the model to other novel (test) logo classes,
that is, Zero-Shot Learning (ZSL) \cite{xu2017transductive,lampert2014attribute,frome2013devise}.
This class-to-class model transfer and generalisation in ZSL is
achieved by knowledge sharing through an intermediate semantic representation
for all classes, such as mid-level attributes \cite{lampert2014attribute} or 
a class embedding space of word vectors \cite{frome2013devise}. 
However, they are limited as many logos do not share attributes or other forms
of semantic representations due to their unique
characteristics. 
%
A lack of large scale logo datasets (Table \ref{tab:dataset}), in both class numbers and
image instance numbers per class, limits severely learning scalable logo detection models. 
This study explores the webly data learning principle
for addressing both large scale dataset construction and incremental logo 
detection model learning without exhaustive manual labelling of increasing image data expansion.
We call this setting {\em scalable logo detection}.

The {\bf contributions} of this work are three-fold:
{\bf (1)} We investigate the {scalable} logo detection problem, 
characterised by modelling {a large quantity of logo classes}
{\em without} exhaustive bounding box labelling. 
This is different from existing methods 
typically considering only a small number of logo classes
with the need for exhaustive manual labelling
at the fine-grained object bounding box level for each class.
This scalability problem is under-studied in the literature. 
%
{\bf (2)}
We propose 
a novel incremental learning approach to scalable deep learning logo detection by
exploiting multi-class detection with synthetic context augmentation.
We call this method {\em Scalable Logo Self-co-Learning} (SL$^2$),
since it automatically discovers potential positive logo images from
noisy web data to progressively improve the model discrimination and generalisation capability
in an iterative {\em joint self-learning and co-learning} manner.
%
%
{\bf (3)} We introduce a large logo detection dataset including 2,190,757 images from 194 logo classes, 
called {\em WebLogo-2M}, created by 
{\em automatically} sampling webly logo images from the social media Twitter. 
Importantly, this dataset construction scheme allows to further expand easily the dataset with 
new logo classes and images, therefore offering a favourable solution for 
{\em scalable} dataset construction.
%
Extensive experiments 
demonstrate the superiority of the SL$^2$ method
over not only the state-of-the-art strongly (Faster R-CNN \cite{ren2015faster},
SSD \cite{liu2015ssd}, YOLOv2 \cite{redmon2017yolo9000}) and weakly (WSL \cite{Huang-CVPR-2016}) supervised
detection models 
but also webly learning methods (WLOD \cite{chen2015webly}), 
on the newly introduced WebLogo-2M dataset.

\section{Related Works}\label{Related Works}

\noindent {\bf Logo Detection }
Early logo detection methods are established on hand-crafted visual features
(e.g. SIFT and HOG)
and conventional classification models
(e.g. SVM)
\cite{li2014logo,revaud2012correlation,romberg2013bundle,boia2014local,kalantidis2011scalable}.
These methods were only evaluated by small logo datasets with 
a limited number of both logo images and classes.
A few deep methods \cite{iandola2015deeplogo,hoi2015logo,su2016deep,liao2017mutual} have been recently proposed
by exploiting the state-of-the-art object detection models
such as R-CNN \cite{girshick2014rich,ren2015faster,girshick2015fast}.
This in turn inspires 
large data construction \cite{hoi2015logo}.
However, all these existing models are not scalable to real world deployments
due to two stringent requirements: 
(1) Accurately labelled training data per logo class;
(2) Strong object-level bounding box annotations.
This is because, both requirements give rise to 
time-consuming training data collection and annotation,
which is not scalable to a realistically large number of logo classes given limited human labelling effort.
In contrast, our method eliminates both needs by 
allowing the detection model learning from image-level weakly annotated
and noisy images automatically collected from the social media
(webly). As such, we enable automated introduction of any quantity of new logos
for both dataset construction/expansion and model updating without
the need for exhaustive manual labelling.

\vspace{0.1cm}
\noindent {\bf Logo Datasets }
A number of logo detection benchmarking datasets exist in the literature (Table \ref{tab:dataset}).
All existing datasets are constructed {\em manually} and
typically small in both image number and logo category
thus insufficient for deep learning.
Recently, Hoi et al. \cite{hoi2015logo} attempt to address this small logo dataset problem
by creating a larger LOGO-NET dataset. 
However, this dataset is not publicly accessible. 
%
To address this scalability problem, 
we propose to collect logo images {\em automatically} from the social media.
This brings about two unique benefits:
(1) Weak image level labels can be obtained for free;
(2) We can easily upgrade the dataset by expanding the logo category set and 
collecting new logo images without
human labelling therefore scalable to 
any quantity of logo images and logo categories.  
To our knowledge, this is the first attempt to construct a large scale 
logo dataset by exploiting inherently noisy web data.

\vspace{0.1cm}
\noindent {\bf Model Self-Learning }
Self-training is a special type of incremental learning
wherein the new training data are labelled by the model itself -- 
predicting logo positions and class labels in weakly labelled or unlabelled images 
before converting the most confident predictions into 
the training data \cite{nigam2000analyzing}.
%
%
A similar approach to our model is 
the detection model by Rosenberg et al. \cite{rosenberg2005semi}.
This model also explores the self-training mechanism.
However, this method needs a number of per class strongly and 
accurately labelled training data
at the object instance level to initialise their detection model.
Moreover, it assumes all unlabelled images belong to the target object
categories.
These two assumptions limit severely model effectiveness and scalability 
given webly collected training data without any object bounding box labelling 
whilst with a high ratio of noisy irrelevant images.

%

\vspace{0.1cm}
\noindent {\bf Model Co-Learning }
Model co-learning is a generic meta-learning strategy originally designed 
for semi-supervised learning,
based on two sufficient and yet conditionally independent feature representations
with a single model algorithm
\cite{blum1998combining}.
Later on, co-learning was further developed into the designs of
using different model parameter settings \cite{wang2007analyzing} or 
models \cite{goldman2000enhancing,zhou2010semi,jiang2013hybrid} 
on the same feature representation.
Overall, the key is that both models in co-learning need to be independently
effective but also complementary to each other.
Recently, there are some attempts on semi-supervised classification methods
in the co-learning spirit \cite{xu2016co,batra2017cooperative,gong2016multi,fakeri2015multiview}.
%
Beyond these, we further extend the co-learning concept from semi-supervised learning 
to webly learning for scalable logo detection.
In particular, we unite co-learning and self-learning in
a single detection deep learning framework with 
the capability of incrementally improving the logo detection models.
To our knowledge, this is the first attempt of exploiting such 
a {\em self-co-learning} approach in the logo detection literature.
\section{WebLogo-2M Logo Detection Dataset}
\label{sec:dataset}

We present a scalable method to automatically construct a large logo detection dataset, 
called {\em WebLogo-2M}, 
with 2,190,757 webly images from 194 logo classes (Table \ref{tab:WebLogo-2M}). 

%

\begin{table} [h] %
	\setlength{\tabcolsep}{0.3cm}
	\centering
	\caption{Statistics of the WebLogo-2M dataset.
		Numbers in parentheses: the minimum/median/maximum per class.	
	}
	\vskip -0.1cm
	\label{tab:WebLogo-2M}
	\begin{tabular}{c|c|c|c}
		\hline
		Logos & Raw Images & Filtered Images & Noise Rate (\%) \\ 
		\hline \hline
		194 & 4,941,317 & 2,190,757 
		& Varying \\ \hline
		- & - & (6/2583/179,789) & (25.0/90.2/99.8)  \\
		\hline 
	\end{tabular}
\vspace{-0.3cm}
\end{table}

\subsection{Logo Image Collection and Filtering}

\begin{figure*} []
	\centering
	\subfloat{
		\includegraphics[width=1\linewidth]
		{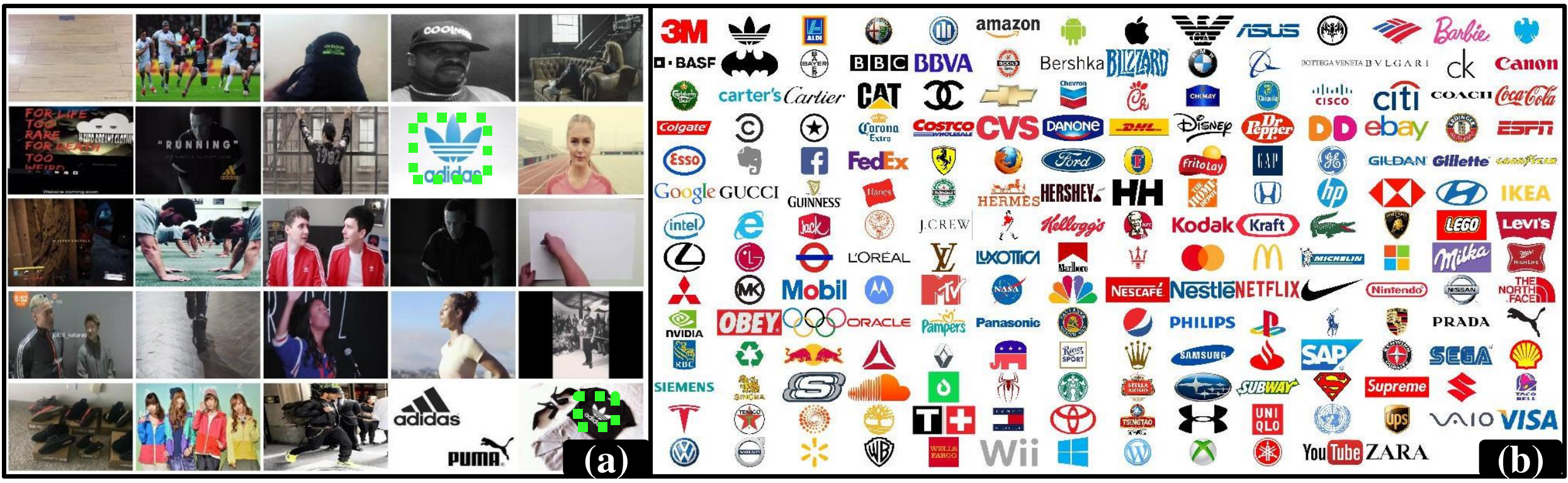}
	}
	\\
	\vskip -0.1cm
	\subfloat{
		\includegraphics[width=1\linewidth]
		{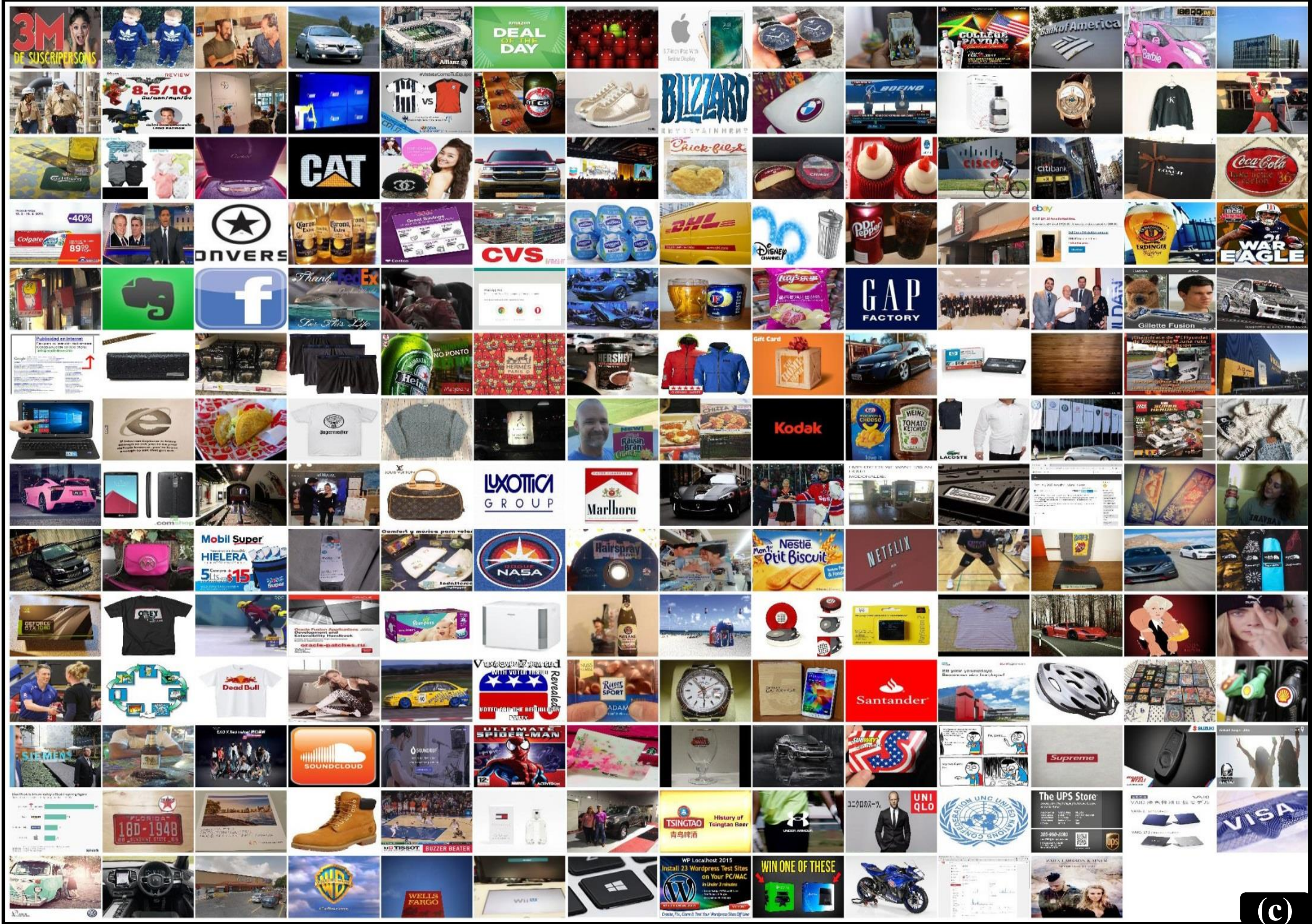}
	}
	\vskip -0.0cm
	\caption{
		A glimpse of the WebLogo-2M dataset.
		{\bf(a)} Example webly (Twitter) logo images randomly selected from the class ``Adidas" 
		with logo instances manually labelled by green dashed bounding boxes only for 
		facilitating viewing.
		Most images contain no ``Adidas" object, i.e. false positives,
		suggesting a high noise degree in webly collected data. 
		{\bf(b)} Clean images of 194 logo classes automatically collected from the Google Image Search, 
		used in synthetic training images generation and augmentation. 
		{\bf(c)} One example true positive webly image per logo class, totally 194 images, showing the rich and diverse context 
		in unconstrained images where typical logo objects
                reside in practice, as compared to those clean logo
                images in (b).
	}
	\label{fig:example_imgs}
\end{figure*}

\noindent {\bf Logo Selection }
A total of 194 logo classes from 13 different categories are selected 
in the WebLogo-2M dataset (Fig. \ref{fig:example_imgs}).
They are popular logos and brands in our daily life, 
including the 32 logo classes of FlickrLogo-32
\cite{romberg2011scalable} and the 10 logo classes of TopLogo-10 \cite{su2016deep}.
%
Specifically, the logo class selection was guided by an extensive review 
of social media reports regarding to the brand popularity 
\footnote{\scriptsize http://www.ranker.com/crowdranked-list/ranking-the-best-logos-in-the-world}\footnote{\scriptsize http://zankrank.com/Ranqings/?currentRanqing=logos}\footnote{\scriptsize http://uk.complex.com/style/2013/03/the-50-most-iconic-brand-logos-of-all-time}
and market-value\footnote{\scriptsize http://www.forbes.com/powerful-brands/list/\#tab:rank}\footnote{\scriptsize http://brandirectory.com/league\_tables/table/apparel-50-2016}.



\vspace{0.1cm}
\noindent {\bf Image Source Selection }
We selected the social media website Twitter 
as the data source of WebLogo-2M.
Twitter offers well structured multi-media data 
stream sources and 
more critically, unlimited data access permission therefore facilitating 
the collection of large scale logo images\footnote{We also attempted at Google and Bing search engines, and three other
social media (Facebook, Instagram, and Flickr). 
However, all of them are rather restricted in data access and limiting
incremental big data collection, e.g. Instagram allows only 500 times of image downloading per hour 
through the official web API.}.

\vspace{0.1cm}
\noindent {\bf Image Collection }
We collected 4,941,317 webly logo images.
Specifically, through the Twitter API, one can automatically retrieve
images from tweets by matching query keywords against 
tweets in real time. 
In our case, we query the logo brand names
so that images in tweets containing the query words can be extracted. 
The retrieved images are then labelled with the corresponding logo name
at the image level, i.e. {\em weakly labelled}.
\vspace{0.1cm}
\noindent {\bf Logo Image Filtering }
We obtained a total of 2,190,757 images
after conducting a two-steps auto-filtering: 
%
%
%
(1) {\em Noise Removal}: 
We removed images of small width and/or height (e.g. less than 100 pixels),
statistically we observed that such images are mostly without
any logo objects (noise). 
%
%
%
(2) {\em Duplicate Removal}: 
We identified and discarded exact-duplicates (i.e. multiple copies of the same image). 
Specifically, given an reference image, 
we removed those with identical width and height.
This image spacial size based scheme is not only computationally cheaper than 
the appearance based alternative \cite{parkhi2015deep},
but also very effective.
For example, we manually examined the de-duplicating process 
on $50$ randomly selected reference images 
and found that over 90$\%$ of the images are true duplicates. 

\subsection{Properties of WebLogo-2M}
\label{sec:dataset_property}

Compared to existing logo detection databases \cite{joly2009logo,romberg2011scalable,hoi2015logo,su2016deep}, 
this webly logo image dataset presents three unique properties
inherent to large scale data exploration for learning scalable logo models:

\vspace{0.1cm}
\noindent {\bf (I) Weak Annotation }
All WebLogo-2M images are weakly labelled at the image level 
by the query keywords. 
These labels are obtained automatically in data 
collection without human fine-grained labelling.
This is much more scalable than manually annotating accurate
individual logo bounding boxes,
particularly when the number of both logo images and classes are very large.

\begin{figure} [h]
	\centering
	\includegraphics[width=1\linewidth]{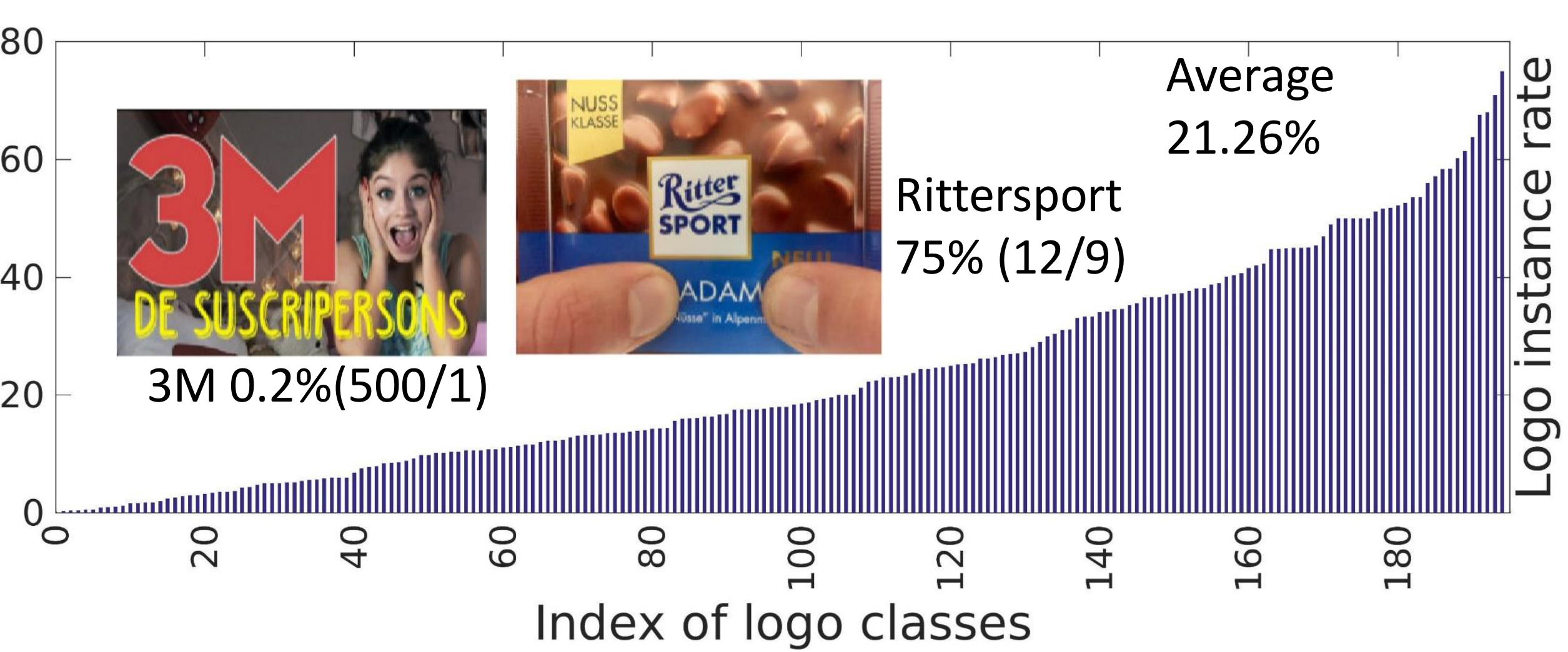}
	\vskip -0.0cm
	\caption{
		True logo image ratios (\%). 
		This was estimated from at most 1,000 random logo images
                per class over 194 classes.
	}
	\label{fig:noisy rate}
	\vspace{-0.1cm}
\end{figure}

\vspace{0.1cm}
\noindent {\bf (II) Noisy (False Positives)}
Images collected from online web sources are inherently noisy,
e.g. often no logo objects appearing in the images therefore providing
plenty of natural false positive samples. 
For estimating a degree of noisiness, 
we sampled randomly at most 1,000 web images per class for all 194 classes and 
manually examined whether they are true or false logo images\footnote{
In the case of sparse logo classes with less than 1,000 webly collected images,
we examined all available images.}.
As shown in Fig. \ref{fig:noisy rate}, the true logo image ratio
varies significantly among 194 logos,
e.g. {\bf 75\%} for ``Rittersport'' vs. {\bf 0.2\%} for ``3M''.
On average, true logo images take only $21.26\%$ vs. 
the remaining as false positives.
Such noisy images pose significant challenges to model learning,
even though there are plenty of data scalable to very large
size in both class numbers and samples per class.




\begin{figure} [h]
	\centering
\includegraphics[width=1\linewidth]{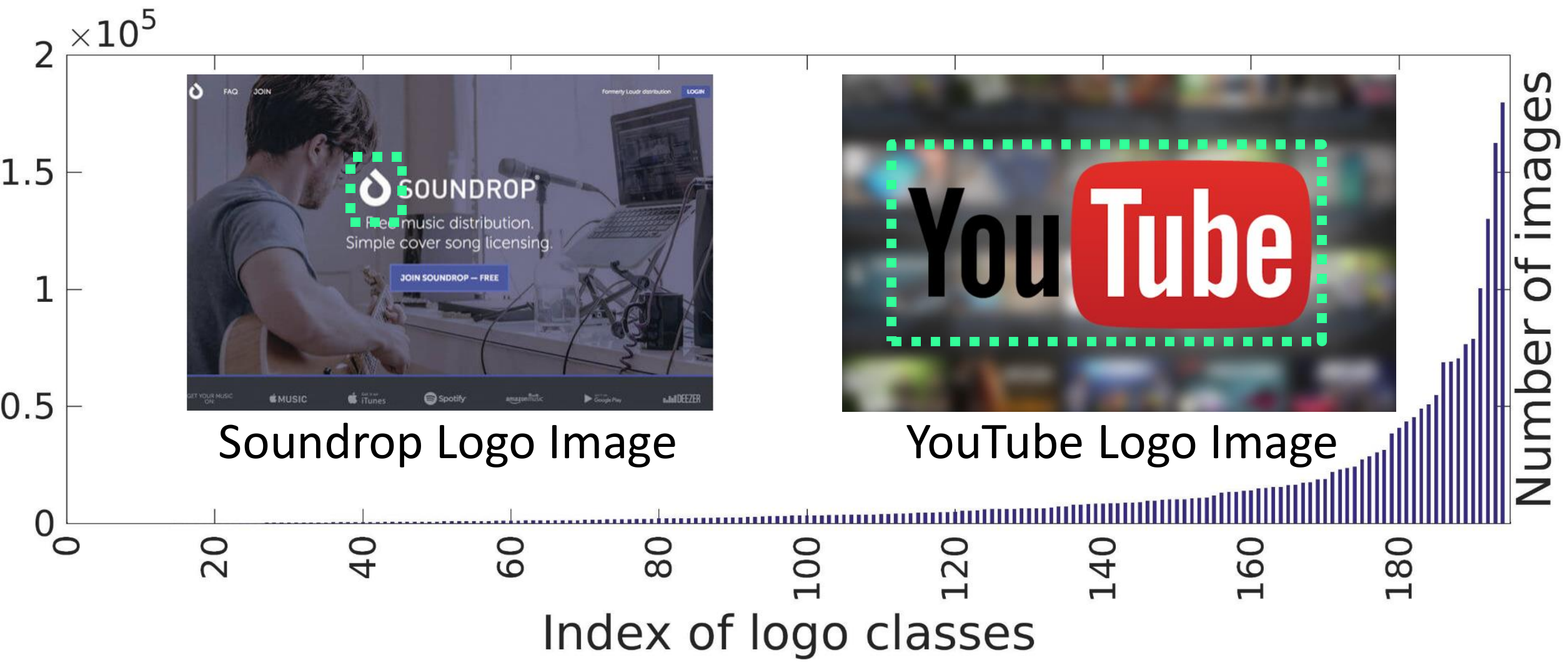}
\vskip -0.0cm
	\caption{
		Imbalanced logo image class distribution, ranging from 6 images (``Soundrop'')
		to 179,789 images (``Youtube''), i.e. 29,965 imbalance ratio.
	}
	\label{fig:classamount}
	\vspace{-0.1cm}
\end{figure}

\vspace{0.1cm}
\noindent {\bf (III) Class Imbalance }
The WebLogo-2M dataset presents a natural logo object occurrence imbalance in public scenes.
Specifically, logo images collected from web streams exhibit a power-law distribution (Fig. \ref{fig:classamount}).
This property is often artificially eliminated in most existing
logo datasets by careful manual filtering,
which not only requires extra labelling effort 
but also renders 
the model learning challenges {\em unrealistic}.
We preserve the inherent class imbalance nature in the data
for achieving fully automated dataset construction and 
retaining realistic model learning challenges.
This requires minimising model learning bias towards
densely-sampled classes \cite{he2009learning}.

\noindent {\bf Further Remark } 
Since the proposed dataset construction method is completely automated,
new logo classes can be easily added without human labelling.
This permits good scalability to enlarging the dataset cumulatively,
in contrast to existing methods \cite{russakovsky2015imagenet,hoi2015logo,lin2014microsoft,everingham2015pascal,joly2009logo,romberg2011scalable,hoi2015logo,su2016deep}
that require exhaustive human labelling therefore hampering further dataset updating and enlarging.
This automation is particularly important
for creating object detection datasets with expensive needs for labelling explicitly object
bounding boxes, more so than for constructing cheaper image-level class annotation datasets \cite{hoffman2014lsda}.
While being more scalable, this WebLogo-2M dataset also provides more
realistic challenges for model learning given weaker label information, noisy image data,
unknown scene context, and significant class imbalance.


\subsection{Benchmarking Training and Test Data}
We define a benchmarking logo detection setting here.
In the scalable webly learning context, 
we deploy the whole WebLogo-2M dataset (2,190,757 images) as
the {\em training} data. 
For performance evaluation, 
a set of images with fine-grained object-level annotation groundtruth is required.
To that end, we construct an independent {\em test set} of 6,558 logo images with logo bounding box labels by
(1) assembling 2,870 labelled images from 
the FlickrLogo-32 \cite{romberg2011scalable} and 
TopLogo \cite{su2016deep} datasets
and (2) manually labelling 3,688 images independently collected from the Twitter.
Note that, the only purpose of labelling this test set is
for performance evaluation of different detection methods,
independent of WebLogo-2M auto-construction.
\section{Training A Multi-Class Logo Detector}







We aim to automatically train a multi-class logo detection model
incrementally
from noisy and weakly labelled web images.
Different from existing methods building a detector in a 
one-pass ``batch'' learning procedure,
we propose to incrementally enhance the model capability 
``sequentially'',
in a joint spirit of self-learning \cite{nigam2000analyzing} and co-learning \cite{blum1998combining}.
This is due to the {\em unavailability} of 
sufficient accurate fine-grained training images per logo class.
In other words, the model must self-select trustworthy
images from the noisy webly labelled data (WebLogo-2M) to 
progressively develop and refine itself.
This is a catch-22 problem: 
The lack of sufficient good-quality training data leads to a suboptimal model
which in turn produces error-prone predictions.
This may cause {\em model drift} -- 
the errors in model prediction will be propagated 
through the iterations therefore 
have the potential to corrupt the model knowledge structure.
Also, the inherent data imbalance over different logo classes
may make model learning biased towards only a few number of
majority classes, therefore leading to significantly
weaker capability in detecting minority classes.
Moreover, the two problems above are intrinsically interdependent 
with possibly one negatively affecting the other.
It is non-trivial to solve these challenges
without exhaustive fine-grained human annotations.
%


\vspace{0.1cm}
\noindent {\bf Formulation Rational } 
In this work, we present a scalable logo detection deep learning solution 
capable of addressing the aforementioned two issues 
in a self-co-learning manner.
The intuition is: 
Web knowledge provides ambiguous but still useful coarse image level logo annotations,
whilst self-learning offers a scalable learning means to explore
iteratively such weak/noisy information
and co-learning allows for mining the complementary advantages of
different modelling approaches in order to further improve
the self-learning effectiveness.
We call our method {\em Scalable Logo Self-co-Learning} (SL$^2$).

\begin{figure*}
	\centering
	\includegraphics[width=1\linewidth]{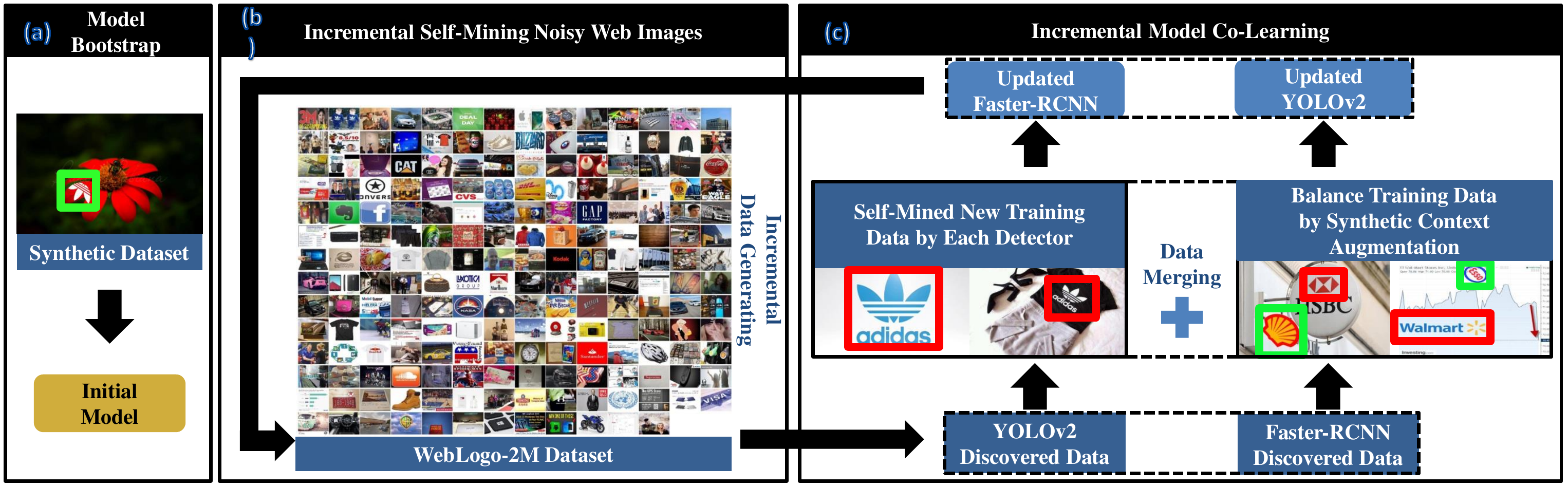}
	\vskip -0.0cm
	\caption{
		Overview of the Scalable Logo Self-co-Learning (SL$^2$) method.
		{\bf(a)} Model initialisation
		by using synthetic logo training images (Sec. \ref{sec:init}).
		{\bf(b)} Incrementally self-mining positive logo images from noisy web data pool
		(Sec. \ref{sec:incremental_learn}). 
		{\bf(c)} Incrementally co-learning the detection models by 
		mined web images and context-enhanced synthetic data
		(Sec. \ref{sec:model_update}).
		This process is repeated iteratively for progressive training data mining and model update.  
	}
	\label{fig:pipeline}
\end{figure*}

\vspace{0.1cm}
\noindent {\bf Model Design } 
To establish a more effective SL$^2$ framework, we select strongly-supervised rather than weakly-supervised 
object detection deep learning models 
for two reasons:
%
%
(1) The performance of weakly-supervised models \cite{cinbis2017weakly} are much inferior than that of strongly supervised counterparts;
(2) The noisy webly weak labels may further hamper the effectiveness of weakly supervised learning. 
In our model instantiation, we choose the Faster R-CNN \cite{ren2015faster}
and YOLO \cite{redmon2017yolo9000} models for self-co-learning.
Conceptually, this model selection is independent of the SL$^2$ notion 
and stronger deep learning detector models generally lead to 
a more advanced SL$^2$ solution.
%
A schematic overview of the SL$^2$ framework is depicted in Fig. \ref{fig:pipeline}.

\subsection{Model Bootstrap}
\label{sec:init}
To start the SL$^2$ process, we first provide reasonably discriminative
logo detection co-learning models
with sufficient bootstrapping training data discovery.
Both Faster R-CNN and YOLOv2 need strongly supervised learning
from object-level bounding box annotations to
achieve logo object detection discrimination, which however 
is not available in our scalable webly learning setting.

To address this problem in the scalable logo detection context, we propose to exploit the idea
of synthesising fine-grained training logo images, 
%
therefore maintaining model learning scalability
for accommodating large quantity of logo classes.
In particular, this is achieved by generating synthetic training images
as in \cite{su2016deep}: 
Overlaying {\em logo icon images}
at random locations of non-logo background images 
so that bounding box annotations
can be {\em automatically} and {\em completely} generated.
The logo icon images are automatically collected from the
Google Image Search by querying the corresponding logo class name 
(Fig. \ref{fig:example_imgs} (b)).
The background images can be chosen flexibly, e.g. the non-logo images 
in the FlickrLogo-32 dataset \cite{romberg2011scalable} and
others retrieved by irrelevant query words from web search engines. 
To enhance appearance variations in synthetic logos,
colour and geometric transformation can be applied \cite{su2016deep}.
%
%

\vspace{0.1cm}
\noindent {\em Training Details } 
We synthesised $1000$ training images per logo class, in total 194,000 images.
For learning the Faster R-CNN and YOLOv2 models,
we set the learning rate at $0.0001$ and the learning iterations 
at $6,000$. 
Following \cite{su2016deep}, 
we pre-trained the detector models on ImageNet 1000-class object classification images
\cite{russakovsky2015imagenet} 
for model warmup.

\subsection{Incremental Self-Mining Noisy Web Images}
\label{sec:incremental_learn}

After the logo detector models are discriminatively bootstrapped, 
we proceed to improve their detection capability with
incrementally self-mined positive (likely) logo images from weakly labelled WebLogo-2M data.
To identify the most compatible training images,
we define a selection function using the detection score of up-to-date model:
\begin{equation}
S(\mathcal{M}_t, \bm{x}, y) = S_{\mbox{det}}(y | \mathcal{M}_t, \bm{x}) 
\label{eq:selection}
\end{equation}
where $\mathcal{M}_t$ denotes the $t$-th step detector model (Faster R-CNN or YOLOv2),
and $\bm{x}$ denotes a logo image with the web image-level label 
$y \!\!\in\!\! Y \!\!=\!\! \{1,2,\cdots,m\}$ with $m$ the total logo class number.
$S_{\mbox{det}}(y | \mathcal{M}_t, \bm{x}) \!\! \in\!\! [0, 1]$, indicates the maximal detection score of $\bm{x}$ on the logo class $y$ by model $\mathcal{M}_t$.
For positive logo image selection, we consider a high threshold detection confidence 
(0.9 in our experiments) \cite{yu2015lsun}
for strictly controlling the impact of model detection errors in degrading
the incremental learning benefits.
This new training data discovery process is summarised in Alg. \ref{alg:self-mining}.


With this self-mining process at $t$-th iteration,
we obtain a separate set of updated training data for
Faster R-CNN and YOLOv2,
denoted as $\mathcal{T}_t^f$ and $\mathcal{T}_t^y$
respectively. 
Each of the two sets represents detection performance with some distinct characteristics 
due to the different formulation designs of the two models, 
e.g. Faster R-CNN is based on region proposals
whilst YOLOv2 relies on pre-defined-grid centred regression.
This creates a satisfactory condition (i.e. diverse and independent modelling) for 
cross-model co-learning.

\newcommand\tab[1][0.5cm]{\hspace*{#1}}
\begin{algorithm} 
	\caption{Incremental Self-Mining Noisy Web Images} 
	\label{alg:self-mining}
	
		\textbf{Input:} 
		Current model $\mathcal{M}_{t-1}$,
		Unexplored data $\mathcal{D}_{t-1}$,
		Self-discovered logo training data $\mathcal{T}_{t-1}$ ($\mathcal{T}_{0} = \emptyset$); \\ [0.15cm]
		\textbf{Output:} 
		Updated self-discovered training data $\mathcal{T}_t$, 
		Updated unlabelled data pool $\mathcal{D}_{t}$; \\ [0.15cm]
		\textbf{Initialisation:} \\
		$\mathcal{T}_t = \mathcal{T}_{t-1}$; \\
		$\mathcal{D}_t = \mathcal{D}_{t-1}$; \\ [0.15cm]
		\textbf{for} image $i$ in $\mathcal{D}_{t-1}$ \\
		\tab Apply $\mathcal{M}_{t-1}$ to get the detection results; \\
		\tab Evaluate image $i$ as a potential positive logo image; \\
		\tab \tab \textbf{if} Meeting selection criterion  \\
		\tab \tab \tab $\mathcal{T}_t$ = $\mathcal{T}_{t} \cup \{i\}$; \\
		\tab \tab \tab $\mathcal{D}_t$ = $\mathcal{D}_{t} \setminus \{i\}$; \\
		\tab \tab\textbf{end if} \\
		\textbf{end for}\\ [0.15cm]
		\textbf{Return} $\mathcal{T}_t$ and $\mathcal{D}_t$.
\end{algorithm}

\subsection{Incremental Model Co-Learning}
\label{sec:model_update}
Given the two up-to-date training sets $\mathcal{T}_t^f$ and $\mathcal{T}_t^y$,
we conduct co-learning on the two detection models (Fig. \ref{fig:pipeline}(d)).
Specifically, we incrementally update the Faster R-CNN model
with the self-mined set $\mathcal{T}_t^y$ by YOLOv2, and 
vice verse. As such, the complementary modelling advantages can be exploited 
incrementally in a self-mining cross-updating manner.

Recall that the logo images across classes are imbalanced (Fig. \ref{fig:classamount}).
This can lead to biased model learning towards well-sampled classes (the {majority classes}), resulting in
poor performance against sparsely-labelled classes (the {minority
	classes}) \cite{he2009learning}.
To address this problem,
we propose the idea of cross-class context augmentation for not only fully exploring the contextual
richness of WebLogo-2M data but also addressing the intrinsic imbalanced logo class problem,
inspired by the potential of context enhancement \cite{su2016deep}.

\begin{figure} [h]
	\centering
	\includegraphics[width=1\linewidth]{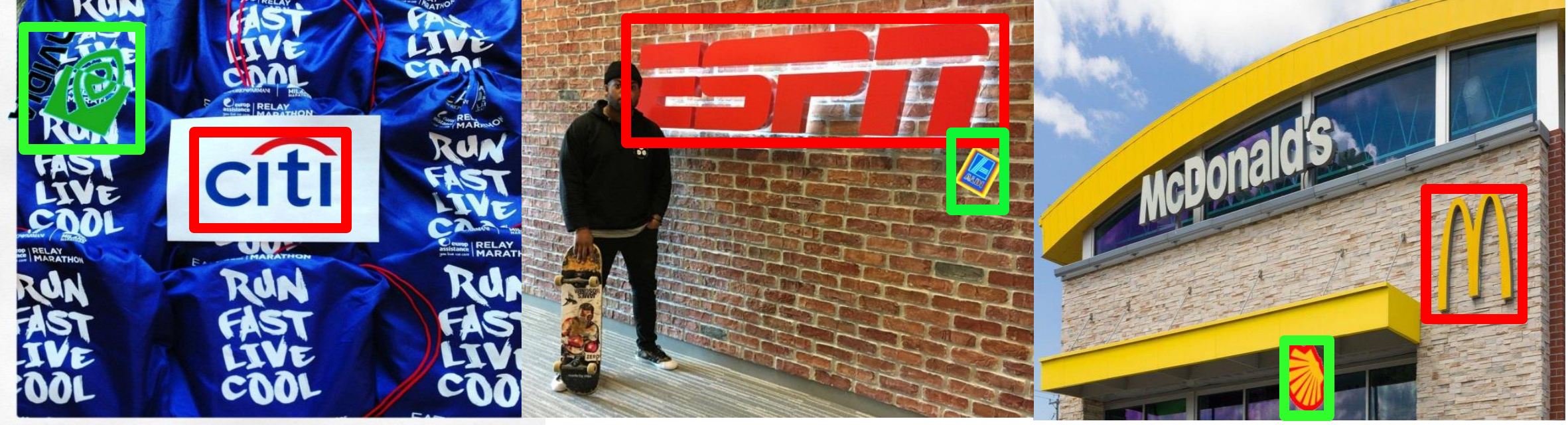}
	\vskip -0.0cm
	\caption{
		Example logo images by synthetic context augmentation. 
		Red box: model detection;
		Green box: synthetic logo ground truth.
	}
	\label{fig:synth_augment}
\end{figure}

Specifically, we ensure that at least $N_\text{cls}$ images will be newly introduced into the 
training data pool in each self-discovery iteration for each detection model. 
Suppose $N_\text{sf}^i$ web images are self-discovered 
for the logo class $i$ (Alg. \ref{alg:self-mining}), 
we generate $N_\text{syn}^i$ synthetic images where
\begin{equation}
N_\text{syn}^i = \max(0, N_\text{cls} - N_\text{sf}^i).
\end{equation}
Therefore, we only perform synthetic data augmentation for those classes
with less than $N_\text{cls}$ real web images mined in the current iteration.
We set $N_\text{cls}=500$ considering that too many synthetic images may bring in
negative effects due to the imperfect logo appearance rendering against background.
Importantly, we choose the self-mined logo images of other classes ($j \neq i$)
as the background images specifically for enriching the contextual diversity
of logo class $i$ (Fig. \ref{fig:synth_augment}).
We utilise the SCL synthesising method \cite{su2016deep} 
as in model bootstrap (Sec. \ref{sec:init}).

Once we have self-mined web training images and context enriched synthetic data,
we perform detection model fine-tuning at the learning rate of $0.0001$ by $6,000$$\sim$$14,000$ iterations 
depending on the training data size at each iteration.
We adopt the original deep learning loss formulation for both Faster R-CNN and YOLOv2.
Model generalisation is improved when
the training data quality is sufficient in terms of both true-false logo image ratio and the context richness.

\subsection{Incremental Learning Stop Criterion}
\label{sec:stop_update}
We conduct the incremental model self-co-learning until 
meeting some stop criterion, for example, the model performance
gain becomes marginal or zero.
We adopt the YOLOv2 as the deployment logo detection model
due to its superior efficiency and accuracy (see Table \ref{tab:iter co-training}). 
In practice, we can assess the model performance on an independent set, e.g. the test evaluation data.


%

\section{Experiments}
\label{sec:exp}

\noindent {\bf Competitors } 
We compared the proposed SL$^2$ model with 
six state-of-the-art alternative detection approaches:\\
{\bf (1)} Faster R-CNN \cite{ren2015faster}:
A competitive region proposal driven object detection model which is
characterised by jointly learning region proposal generation and
object classification in a single deep model.
In our scalable webly learning context, the Faster R-CNN is 
optimised with synthetic training data
generated by the SCL \cite{su2016deep} method, 
exactly the same as our SL$^2$ model.\\
%
{\bf (2)} SSD \cite{liu2015ssd}:
A state-of-the-art regression optimisation based object detection model.
We similarly learn this strongly supervised model with
synthetic logo instance bounding box labels as Faster R-CNN above. \\
%
%
{\bf (3)} YOLOv2 \cite{redmon2017yolo9000}:
A contemporary bounding box regression based
multi-class object detection model.
We learned this model with the same training data as SSD and Faster R-CNN. 
\\
{\bf (4)} Weakly Supervised object Localisation (WSL) \cite{Huang-CVPR-2016}:
A state-of-the-art weakly supervised detection model allowing to be trained with 
image-level logo label annotations in a multi-instance learning framework.
Therefore, we can directly utilise the webly labelled WebLogo-2M images 
to train the WSL detection model. Note that, noisy logo labels inherent to web data
may pose additional challenges in addition to high complexity in logo appearance and context. \\
%
{\bf (5)} Webly Learning Object Detection (WLOD) \cite{chen2015webly}: 
A state-of-the-art weakly supervised object detection method where
clean Google images are used to train exemplar classifiers which 
is deployed to classify region proposals by EdgeBox \cite{ZitnickECCV14edgeBoxes}. 
In our implementation, we further improved the classification component 
by exploiting an ImageNet-1K and pascalVOC trained VGG-16 \cite{simonyan2014very} model as the deep feature extractor and the L2 distance as the matching metric.
We adopted the nearest neighbour classification model with 
Google logo images (Fig. \ref{fig:example_imgs}(b))
as the labelled training data. \\
%
{\bf (6)} WLOD+SCL: 
a variant of WLOD \cite{chen2015webly} with context enriched training data by
exploiting SCL \cite{su2016deep} to synthesise various context for 
Google logo images.\\
Overall, these existing methods
cover 
both {\em strongly} and {\em weakly} supervised learning
based detection models.



\vspace{0.1cm}
\noindent {\bf Performance Metrics }
For the quantitative performance measure of logo detection, 
we utilised the Average Precision (AP) for each individual logo class,
and the mean Average Precision (mAP) for all classes \cite{everingham2010pascal}.
A detection is considered being correct when the Intersection over Union (IoU) between the predicted and groundtruth exceeds $50\%$.

\subsection{Comparative Evaluations}

\begin{table}  
	\centering
	\caption{Logo detection performance on WebLogo-2M.}
	\setlength{\tabcolsep}{0.7cm}
	\vskip -0.2cm
	\label{tab:main}
	\begin{tabular}{c|c}
		\hline
		Method & mAP (\%) \\
		\hline \hline
		SSD \cite{liu2015ssd} 
		&  8.8 \\
		\hline 
		Faster R-CNN \cite{ren2015faster}
		& 14.9 \\ \hline
		YOLOv2 \cite{redmon2017yolo9000} 
		& 18.4 \\ \hline
		\hline
		
		WSL \cite{Huang-CVPR-2016} 
		& 3.6 \\
		
		\hline
		WLOD \cite{chen2015webly} & 19.3 \\
		
		WLOD\cite{chen2015webly} + SCL\cite{su2016deep} & 7.8 \\

		\hline \hline
		\bf SL$^2$ (Ours)
		& \bf 46.9 \\
		\hline
	\end{tabular}
	\vspace{-0.0cm}
\end{table}


%
%
%
%
%
%

\begin{figure} 
	\centering
	\includegraphics[width=1.0\linewidth]{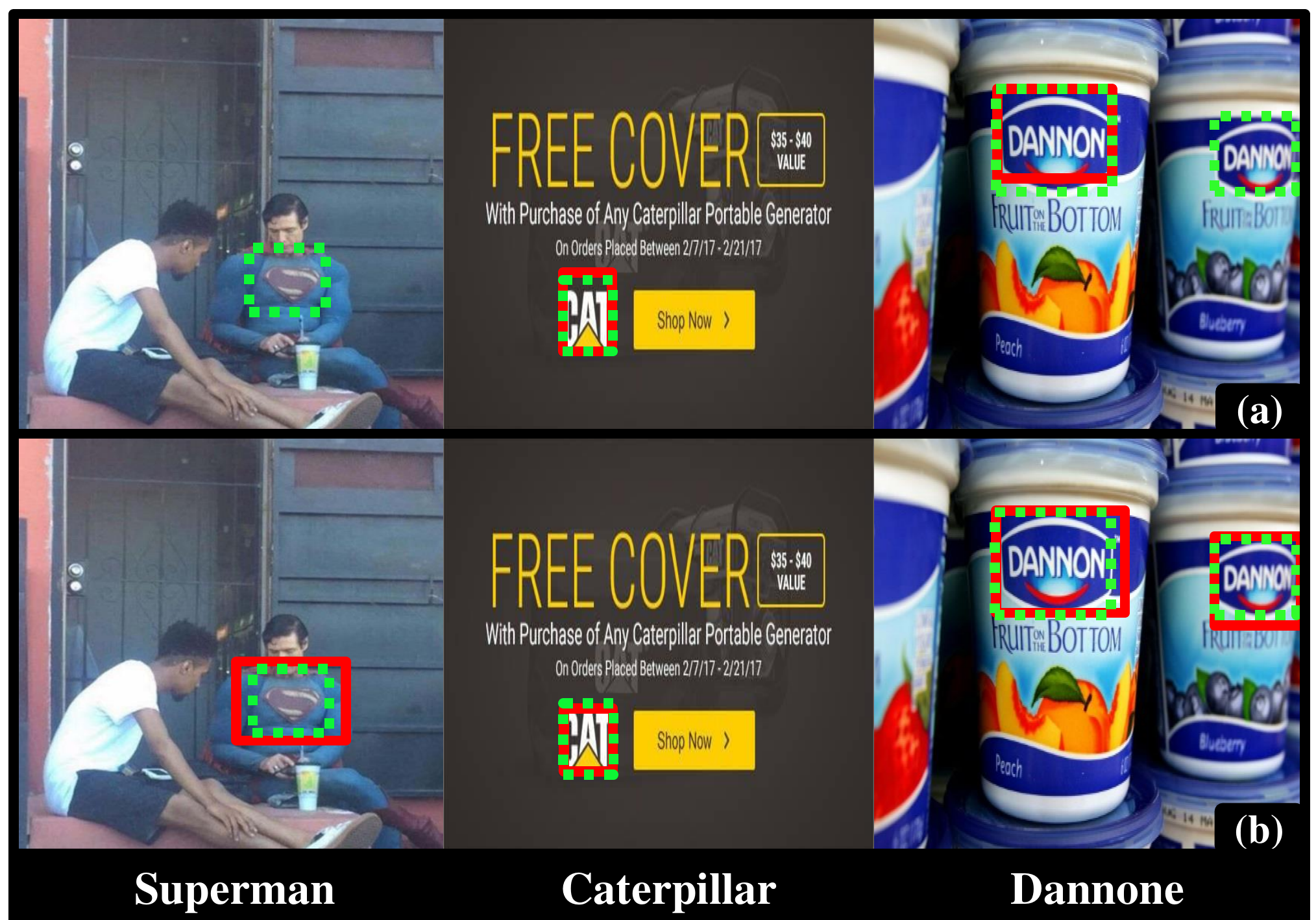}
	\vskip -0.0cm
	\caption{
		{Quantitative evaluations of the {\bf (a)} WLOD and {\bf (b)} SL$^2$ models. 
			Green dashed boxes: ground truth. Red solid boxes: detected. 
			WLOD fails to detect visually ambiguous ($1^\text{st}$ column) logo instance, success on relatively clean ($2^\text{nd}$ column) logo instances, while only fires partially on the salient one ($3^\text{rd}$ column). 
			The SL$^2$ model can correctly detect all these logo instances with varying context and appearance quality.}
	}
	\label{fig:det_samp}
\end{figure}

\begin{table*} [t]  
	\centering
	\setlength{\tabcolsep}{0.45cm}
	\caption{Model performance development over incremental SL$^2$ iterations.}
	\vskip -0.2cm
	\label{tab:iter_mAP}
	\begin{tabular}{c||c|c|c|c|c|c|c|c|c}
		\hline 
		Iteration & 0 & 1$^\text{st}$ & 2$^\text{nd}$ & 3$^\text{rd}$ & 4$^\text{th}$ & 5$^\text{th}$ & 6$^\text{th}$ & 7$^\text{th}$ & 8$^\text{th}$   \\ \hline \hline
		mAP (\%) & 18.4 & 28.6 & 33.2 & 39.1 & 42.2 & 44.4 & 45.6 &\bf 46.9 & \bf 46.9 
		\\
		\hline
		mAP Gain (\%) & N/A & \bf 10.2 & 4.6 & 5.9 & 3.1 & 2.2 & 1.2 & 1.3 & 0.0 
		\\
		\hline 
		Training Images & 5,862 & 21,610 & 41,314 & 54,387 & 74,855 & 86,599 & 98,055 &\bf 107,327 &\color{blue} \bf Stop 
		\\
		\hline
	\end{tabular}
\end{table*}

We compared the scalable logo detection performance 
on the WebLogo-2M benchmarking test data in Table \ref{tab:main}.
It is evident that the proposed SL$^2$ model significantly outperforms all other alternative methods,
e.g. surpassing the best baseline WLOD by 27.6\% (46.9\%-19.3\%) in mAP.
We also have the following observations:
\\
{\bf (1)} The weakly supervised learning based model WSL produces the worst result,
due to the joint effects of complex logo appearance variation against unconstrained context and
the large proportions of false positive logo images (Fig. \ref{fig:noisy rate}).
\\
{\bf (2)} The WLOD method performs reasonably well suggesting that 
the knowledge learned from auxiliary data sources (ImageNet and Pascal VOC) 
is transferable to some degree, confirming the similar findings as in \cite{sharif2014cnn,yosinski2014transferable}. 
\\
{\bf (3)} By utilising synthetic training images with rich context and background, 
fully supervised detection models YOLOv2 and
Faster R-CNN are able to achieve the 3$^\text{rd}$/$4^\text{th}$ best results among all competitors.
This suggests that context augmentation is critical for object detection model optimisation, and the combination of {\em strongly} supervised learning model + auto training data synthesising
is a superior strategy over {\em weakly} supervised learning in webly learning setting.
\\
{\bf (4)}
Another supervised model SSD
yields much lower detection performance. 
This is similar to the original finding
that this model is sensitive to the bounding box size of objects 
with weaker detection performance on small objects such as 
in-the-wild logo instances.
\\
{\bf (5)} WLOD+SCL produces a weaker result (7.8\%) compared to WLOD (19.3\%).
This indicates that joint supervised
learning is critical to exploit context enriched data augmentation,
otherwise likely introducing some distracting effects resulting in degraded matching. \\
%
%

\vspace{0.1cm}
\noindent {\em Qualitative Evaluation } 
For visual comparison, we show a number of qualitative 
logo detection examples from three classes by the SL$^2$ and WLOD models Fig. \ref{fig:det_samp}.

\subsection{Further Analysis and Discussions}

\subsubsection{Effects of Incremental Model Self-Co-Learning}
\begin{figure} 
	\centering
	\includegraphics[width=1\linewidth]{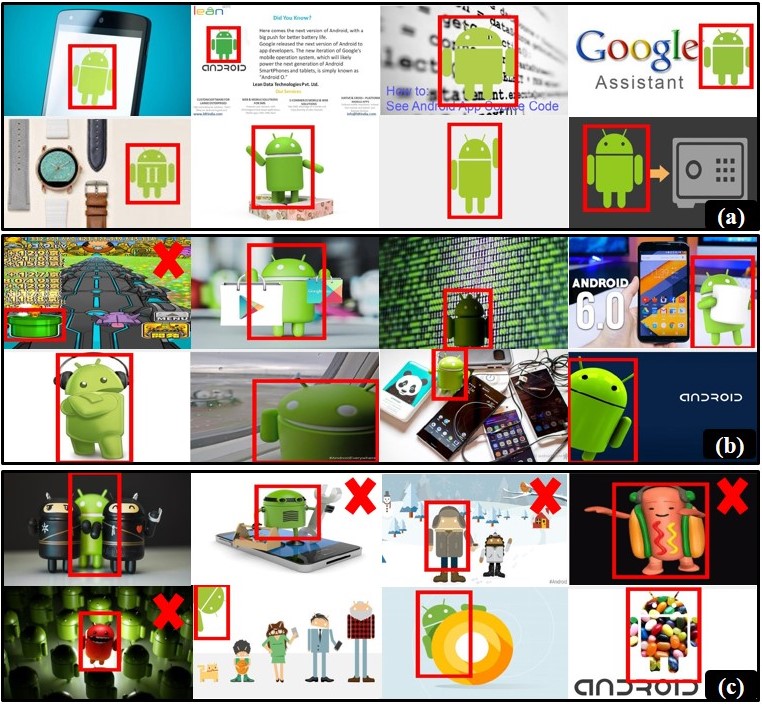}
	\vskip -0.0cm
	\caption{
		Randomly selected images self-discovered in the 
		{\bf (a)} $1^\text{st}$,
		{\bf (b)} $4^\text{th}$, and
		{\bf (c)} $8^\text{th}$ iterations for the logo class ``Android''. 
		Red box: SL$^2$ model detection. 
		Red cross: 
		false detection. 
		The images mined in the $1^\text{st}$ iteration have clean logo instances and background, whilst those discovered in the $4^\text{th}$ and $8^\text{th}$ iterations have more 
		diverse logo appearance variations
		in richer and more complex context.
		More false detections are likely to be produced in the $4^\text{th}$ 
		and $8^\text{th}$ self-discovery.
	}
	\label{fig:iter_imgs}
\end{figure}

\begin{figure} [h]
	\centering
	\includegraphics[width=1.0\linewidth]{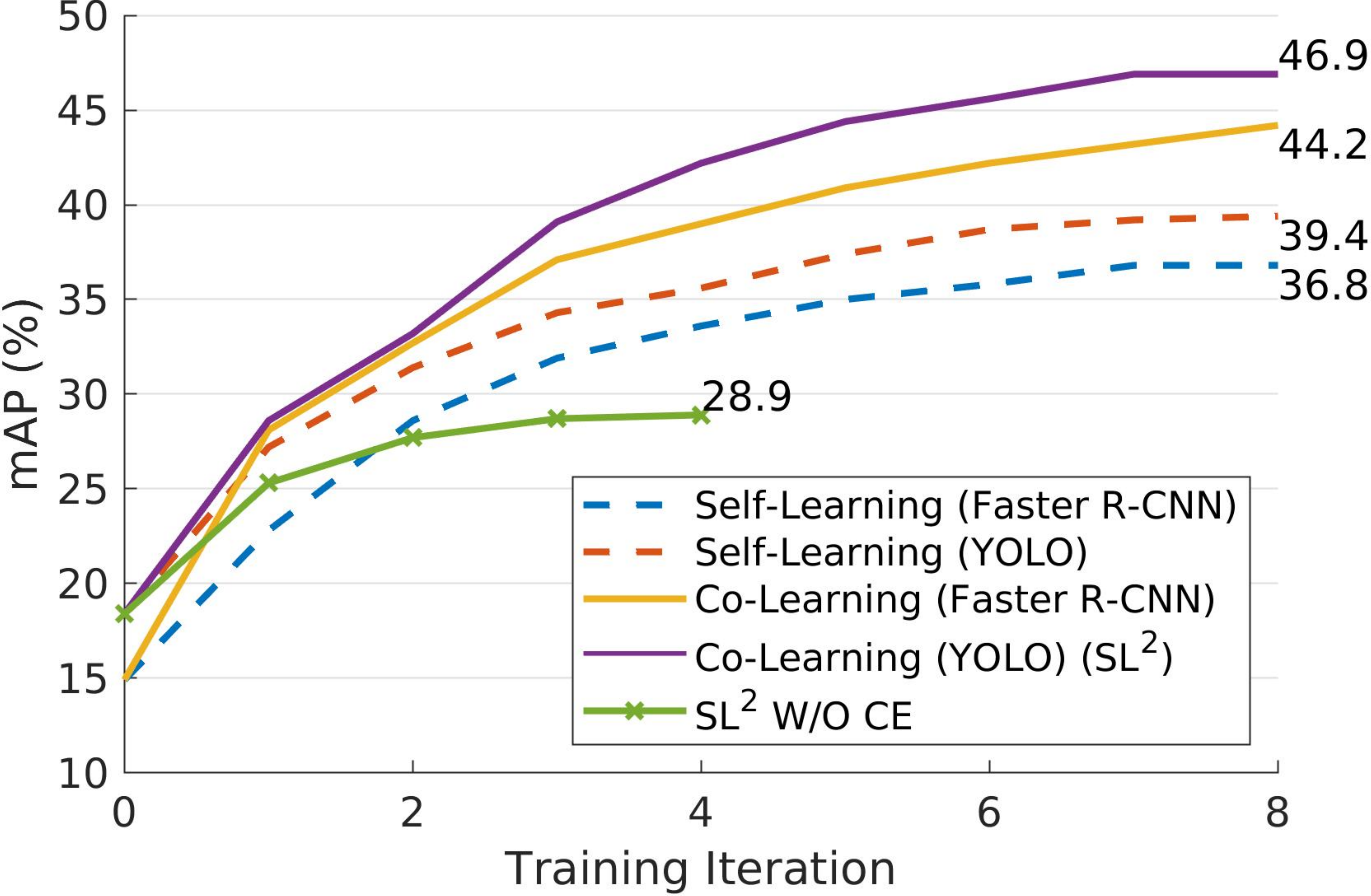} 
	\vskip -0.0cm
	\caption{
		Evaluating the model co-learning and self-learning strategies, and
		the effect of Context Enhancement (CE) based training data class balancing
		in the SL$^2$ on WebLogo-2M.
	}
	\label{fig:increamental graph}
	\vspace{-0.0cm}
\end{figure}

We evaluated the effects of incremental model self-co-learning on self-discovered
training data and context enriched synthetic images
by examining the SL$^2$ model performance at individual iterations.
Table \ref{tab:iter_mAP} and Fig. \ref{fig:increamental graph} show 
that the SL$^2$ model improves consistently from the $1^\text{st}$ to $8^\text{th}$ iterations of self-co-learning.
In particular,
the starting data mining brings about the maximal mAP gain
of 10.2\% (28.6\%-18.4\%) with the per-iteration benefit mostly dropping gradually.
This suggests that our model design is capable of effectively addressing the notorious error propagation challenge thanks to 
(1) a proper detection model initialisation 
by logo context synthesising for providing a sufficiently good starting-point detection;
(2) a strict selection on self-evaluated detections for reducing the amount of false positives, suppressing the likelihood of error propagation;
and 
(3) cross-model co-learning with cross-logo context enriched synthetic training data augmentation 
with the capability of 
addressing the imbalanced data learning problem whilst enhancing the model robustness
against diverse unconstrained background clutters.
We also observed that more images are mined along the incremental
data mining process, suggesting that the SL$^2$ model improves over time
in the capability of tackling more complex context, although potentially  
leading to more false positives simultaneously which can
cause lower model growing rates, as indicated in Fig. \ref{fig:iter_imgs}.


\begin{table} [h] 
	\centering
	\setlength{\tabcolsep}{0.7cm}
	\caption{Co-learning {\em versus} self-learning.}
	\vskip -0.2cm
	\label{tab:iter co-training}
	\begin{tabular}{l|c}
		\hline
		Method & mAP (\%)\\
		\hline \hline
		Self-Learning (Faster R-CNN)
		&  36.8 \\ 
		\hline
		Self-Learning (YOLO)
		&  39.4 \\
		\hline 	\hline
		Co-Learning (Faster R-CNN)
		&  44.2 \\ 
		\hline
		 Co-Learning (YOLO) ({\bf SL$^2$}) 
		&  \bf 46.9 \\ 
		\hline
		
	\end{tabular}
	\vspace{-0.0cm}
\end{table}

\vspace{0.1cm}
\subsubsection{Effects of Cross-Model Co-Learning}
We assessed the benefits of cross-model co-learning between
Faster R-CNN and YOLOv2 in SL$^2$ 
in comparison of the single-model {\em self-learning} strategy.
In contrast to co-learning, 
the self-learning exploits self-mined new training data for incremental model update
without the benefit of cross-model complementary advantages.
Table \ref{tab:iter co-training} and Fig. \ref{fig:increamental graph} show that
both models benefit clear performance gains from co-learning,
e.g. 7.4\% (44.2-36.8) for Faster R-CNN,
and 7.5\% (46.9-39.4) for YOLOv2. 
This verifies the motivation and our idea of 
exploiting the co-learning principle for
maximising the complementary information of 
different detection modelling formulations
in the scalable logo detection model optimisation.

%
%


\begin{table}  
	\centering
	\setlength{\tabcolsep}{0.3cm}
	\caption{Effects of training data Context Enhancement (CE). Metric: mAP (\%).}
	\vskip -0.2cm
	\label{tab:balance}
	\begin{tabular}{c||c|c|c|c|c|c}
		\hline 
		Iteration & 0 & 1$^\text{st}$ & 2$^\text{nd}$ & 3$^\text{rd}$ & 4$^\text{th}$ & 5$^\text{th}$ 
		\\ \hline \hline
		{\bf With} CE & 18.4 &\bf 28.6 &\bf 33.2 &\bf 39.1 &\bf 42.2 &\bf 44.4 
		\\
		\hline
		{\bf Without} CE & 18.4 & 25.3 & 27.7 & 28.7 & 28.9 & 28.0 \\
		\hline
	\end{tabular}
\end{table}


\vspace{0.1cm}
\subsubsection{Effects of Synthetic Context Enhancement}
We evaluated the impact of training data context enhancement (i.e. the cross-class context enriched synthetic training data) on the SL$^2$ model performance.
Table \ref{tab:balance} shows that 
context enhancement not only provides clearly model improvement across iterations
due to the suppression of negative imbalance learning effect and enriched context knowledge,
but also simultaneously enlarges the data mining capacity due to potentially less noisy training data aggregation.
Without context enhancement and training class balancing, the model
stops to improve by the 4$^\text{th}$ iteration of the incremental learning, 
with much weaker model generalisation performance
at 28.9\% vs 46.9\% by the full SL$^2$.
This suggests the importance of context and data balance in detection model learning, 
therefore validating our model design considerations.

\section{Conclusion}
We present a scalable end-to-end logo detection solution 
including logo dataset establishment and multi-class logo detection model learning, 
realised by exploring the webly data learning principle
without
the tedious cost of manually labelling fine-grained logo annotations.
%
Particularly, we propose a new incremental learning method named 
Scalable Logo Self-co-Learning (SL$^2$) for enabling reliable self-discovery 
and auto-labelling of new training images from noisy web data to progressively
improve the model detection capability in a cross-model 
co-learning manner given unconstrained in-the-wild images.
Moreover, we construct a very large logo detection benchmarking dataset WebLogo-2M by automatically collecting and processing web stream data
(Twitter) in a scalable manner,
therefore facilitating and motivating further investigation of 
scalable logo detection in future studies by the wider community. 
We have validated the advantages and superiority of the 
proposed SL$^2$ approach
in comparisons to the state-of-the-art alternative methods ranging from
strongly-supervised and weakly-supervised detection models to webly data learning
models through extensive comparative evaluations and 
analysis on the benefits of incremental model training and context
enhancement, using the newly introduced WebLogo-2M logo benchmark dataset.
We provide in-depth SL$^2$ model component analysis and evaluation
with insights on model performance gain and formulation.

\section*{Acknowledgement}
{\noindent This work was partially supported by the China Scholarship Council, Vision Semantics Ltd., 
	the Royal Society Newton Advanced Fellowship Programme (NA150459), 
and InnovateUK Industrial Challenge Project on Developing and Commercialising Intelligent Video Analytics Solutions for Public Safety.}

\ifCLASSOPTIONcaptionsoff
  \newpage
\fi



%

{\small
\bibliographystyle{IEEEtran}
\bibliography{record}
}

\end{document}